\documentclass{article}

\usepackage{booktabs} % For formal tables
\usepackage{amsmath}
\usepackage{xcolor}
\usepackage{algorithm,algpseudocode}
\usepackage{graphicx}
\usepackage{grffile}
\usepackage{subcaption}
\usepackage{caption}
\usepackage{colortbl}
\usepackage{breqn}
\usepackage{balance}
\usepackage[numbers,sort&compress,longnamesfirst,sectionbib]{natbib}
\usepackage{authblk}

\newcommand{\cellbluelight}{{\cellcolor{blue!15}}}

\newcommand{\ignore}[1]{}

\newcommand{\overbar}[1]{\mkern 1.5mu\overline{\mkern-1.5mu#1\mkern-1.5mu}\mkern 1.5mu}

\newtheorem{definition}{Definition}

\newcommand{\TTP}{TTP}
\newcommand{\TSP}{TSP}
\newcommand{\KP}{KP}
\newcommand{\BOTTP}{BO-TTP}
\newcommand{\DP}{DP}
\newcommand{\PWT}{PWT}

\begin{document}
\title{Evolutionary Computation plus Dynamic Programming for the Bi-Objective Travelling Thief Problem}
\date{}

\author[1]{Junhua Wu}
\author[2]{Sergey Polyakovskiy}
\author[1]{Markus Wagner}
\author[1]{Frank Neumann}
\affil[1]{School of Computer Science, The University of Adelaide}
\affil[2]{School of Information Technology, Deakin University}

\maketitle

\begin{abstract}
This research proposes a novel indicator-based hybrid evolutionary approach that combines approximate and exact algorithms. We apply it to a new bi-criteria formulation of the travelling thief problem, which is known to the Evolutionary Computation community as a benchmark multi-component optimisation problem that interconnects two classical $\mathcal{NP}$-hard problems: the travelling salesman problem and the 0-1 knapsack problem. Our approach employs the exact dynamic programming algorithm for the underlying Packing-While-Travelling problem as a subroutine within a bi-objective evolutionary algorithm. This design takes advantage of the data extracted from Pareto fronts generated by the dynamic program to achieve better solutions. Furthermore, we develop a number of novel indicators and selection mechanisms to strengthen synergy of the two algorithmic components of our approach. The results of computational experiments show that the approach is capable to outperform the state-of-the-art results for the single-objective case of the problem.
\end{abstract}

%%%%%%%%%%%%%%%%%%%%%%%%%%%%%%%%%%%%%%%%%%%%%%%%%%%%%%%%%%%%%%%%%%%%%%%%%%%%%%%%%%%%%%%%%%%%%%%%%%%%%%%%%%%%%%%%%%
\section{Introduction}
%%%%%%%%%%%%%%%%%%%%%%%%%%%%%%%%%%%%%%%%%%%%%%%%%%%%%%%%%%%%%%%%%%%%%%%%%%%%%%%%%%%%%%%%%%%%%%%%%%%%%%%%%%%%%%%%%%
The travelling thief problem (TTP)~\cite{DBLP:conf/cec/BonyadiMB13} is a bi-component problem, where two well-known $\mathcal{NP}$-hard combinatorial optimisation problems, namely the travelling salesperson problem ({\TSP}) and the 0-1 knapsack problem ({\KP}), are interrelated. Hence, tackling each component individually is unlikely to lead to a global optimal solution. It is an artificial benchmark problem modelling features of complex real-world applications emerging in the areas of planning, scheduling and routing. For example, \citet{DBLP:journals/ori/StolkMMM13} exemplify a delivery problem that consist of a routing part for the vehicle(s) and a packing part of the goods onto the vehicle(s).

Thus far, many approaches have been proposed for the TTP \cite{DBLP:conf/gecco/PolyakovskiyB0MN14, Bonyadi2014gecco, DBLP:conf/seal/MeiLY14, DBLP:conf/cec/MeiLSY15, DBLP:journals/soco/MeiLY16, Faulkner2015, 7321437, Wagner2016, ELYAFRANI2017795, ElYafrani:2016:PVS:2908812.2908847, ElYafrani2018231, Martins:2017:HHS:3071178.3071235, ElYafrani2017, Lourenco2016}. However to the best of our knowledge, all of them are focusing on utilising the existing heuristic approaches (such as local search, simulated annealing, tabu search, genetic algorithms, memetic algorithm, swarm intelligence, etc.), incorporating either well-studied operators of the TSP and KP or slight variations of such operators. The heuristic approaches or operators that take advantage of the existing exact algorithms of the TTP \cite{DBLP:journals/corr/NeumannPSSW17, DBLP:conf/seal/Wu0PN17} are yet lacking. On the other hand, very few investigations have been taken on the approaches of the multi-objective formulations of the TTP except by \citet{Blank2017, Yafrani2017}.

In this paper, we consider a bi-objective version of the TTP, where the goal is to minimise the weight and maximise the overall benefit of a solution. We present a hybrid approach for the bi-objective TTP that uses the dynamic programming approach for the underlying PWT problem as a subroutine. The evolutionary component of our approach constructs a tour $\pi$ for the TTP. This tour is then fed into the dynamic programming algorithm to compute a trade-off front for the bi-objective problem. Here the tour $\pi$ is kept fixed and the resulting packing solutions are Pareto optimal owing to the capability of the dynamic programming. A key aspect of the algorithm is to take advantage of the different fronts belonging to different tours for the TTP component, as presumably the global Pareto optimum might contain some segments from the different fronts. Meanwhile, when the evolutionary approach evolves the tours and the current general Pareto front consists of different tours (together with the packing plans), a challenge is to select tours for mutations and crossovers that lead to promising new tours. Such tours shall result in new Pareto optimal solutions for the overall bi-objective TTP problem when running the dynamic programming on them. In short, the selection mechanism shall encourage the synergy of the two sub-approaches. We introduce a novel indicator-based evolutionary algorithm (IBEA \cite{DBLP:conf/ppsn/ZitzlerK04}) that contains a series of customised indicators and parent selections to achieve this goal. Our results show that this approach solves the problem well, and its by-product, which is the total reward of the single objective TTP, beats the state-of-the-art approach in most cases.

The remainder of the paper first states the bi-objective version of the {\TTP} mathematically in Section~\ref{sec:ttp}. Then, Section~\ref{sec:prelim} covers the prerequisites required for our approach, which is later introduced in Section~\ref{sec:approach}. Section~\ref{sec:experiments} provides the description of the computational setup and the analysis of computational experiments. Finally, Section~\ref{sec:conclusion} draws conclusions.

%%%%%%%%%%%%%%%%%%%%%%%%%%%%%%%%%%%%%%%%%%%%%%%%%%%%%%%%%%%%%%%%%%%%%%%%%%%%%%%%%%%%%%%%%%%%%%%%%%%%%%%%%%%%%%%%%%
\section{The Travelling Thief Problem}\label{sec:ttp}
%%%%%%%%%%%%%%%%%%%%%%%%%%%%%%%%%%%%%%%%%%%%%%%%%%%%%%%%%%%%%%%%%%%%%%%%%%%%%%%%%%%%%%%%%%%%%%%%%%%%%%%%%%%%%%%%%%

The standard single-objective {\TTP} \cite{DBLP:conf/gecco/PolyakovskiyB0MN14} involves $n$ cities, $m$ items, and a thief who must make a tour visiting each of the cities exactly once. The cities form a set of nodes $V=\left\{1,\ldots,n\right\}$ in a complete graph $G=\left(V,E\right)$, where $E\subseteq V^2$ is a set of edges representing all possible connections between the cities. Every edge $e_{ij}\in E$ is assigned a known distance $d_{ij}$. Every node $i\in V$ but the first one relates to a unique set of items $M_i=\{1,\ldots,m_i\}$, $\textstyle\sum_{i=2}^n{m_i}=m$, stored in the corresponding city. Each item $k\in M_i$ positioned in node $i$ is associated with an integer profit $p_{ik}$ and an integer weight $w_{ik}$. The thief starts and ends the tour in the first node and can collect any of the items located in the intermediate nodes $2,\ldots,n$. Items may only be selected until their total weight exceeds a knapsack's capacity $C$. Furthermore, the thief pays a rent rate $R$ for each time unit of travelling. Selection of an item contributes its profit to a total reward, but produces a transportation cost relative to its weight. As the weight of each added item slows down the thief, the transportation cost increases. This cost is therefore deducted from the reward. When the knapsack is empty, the thief can achieve a maximal velocity $\upsilon_{max}$. When it is full, the thief can only move with a minimal velocity $\upsilon_{min}>0$. The actual velocity $\upsilon_i$ when moving along the edge $e_{ij}$ depends on the total weight of items chosen in the cities preceding $i$. The problem asks to determine a combination of a tour and a subset of items that minimises the difference between the total profit of selected items and the overall transportation cost. 

Let an integer-valued vector $\pi \in V^n$, $\pi=(\pi_1,...,\pi_n)$, represent a tour such  that $\pi_i=j$ iff $j$ is the $i$th visited node of the tour. Clearly, $\pi_i \neq \pi_j$ for any $i,j\in V$, $i\neq j$. Next, let a binary decision vector $\rho\in \left\{0,1\right\}^m$, $\rho=(\rho_{21},...,\rho_{nm_n})$, encode a packing plan of the problem such that $\rho_{ik}=1$ iff item $k$ in node $i$ is chosen, and $0$ otherwise. Then $W_{\pi_i}=\sum_{j=1}^i{\sum_{k=1}^{m_j} w_{jk} \rho_{jk}}$ is a total weight of items sequentially selected in the nodes from $\pi_1$ to $\pi_i$, and $\upsilon_{\pi_i}=\upsilon_{max}-\nu W_{\pi_i}$, $\nu = \left(\upsilon_{max}-\upsilon_{min}\right)/C$, is the real velocity of the thief quitting the $i$th node. In summary, the objective function of the {\TTP} has the following form:
\begin{flalign}
f\left(\pi,\rho\right) =\!\displaystyle\sum_{i=1}^n \displaystyle\sum_{k=1}^{m_i} p_{ik} \rho_{ik} \!-\! R \left( \displaystyle\sum_{i = 1}^{n-1}\frac{d_{\pi_i \pi_{i+1}}}{\upsilon_{\pi_i}} \!+\! \frac{d_{\pi_n \pi_1}}{\upsilon_{\pi_n}} \right) \label{obj1}
\end{flalign}

Here, we extend the standard formulation of the {\TTP} by introduction of an additional objective function. The new version, named as {\BOTTP} for short, becomes a bi-objective optimisation problem, where the total accumulated weight 
\begin{flalign}
\varphi\left(\rho\right) =\sum_{i=1}^n{\sum_{k=1}^{m_i} w_{ik} \rho_{ik}} \label{obj2}
\end{flalign}
yields the second criterion. Such extension appears natural regarding the {\TTP} as one may either need to maximise the reward for a given weight of collected items, or determine the least weight subject to bounds imposed on the reward. Note that even if $\pi$ is fixed, (\ref{obj1}) is a non-monotone sub-modular function \cite{DBLP:journals/eor/PolyakovskiyN17} that implies possible deterioration of the reward as the number of selected items, and therefore their total weight, increases. 
We formulate the {\BOTTP} as follows:
\begin{flalign}
\nonumber (\pi,\rho) = \!\left\{
\begin{array}{l}
\!\!\arg \max f(\pi, \rho)
\\
\!\!\arg \min \varphi(\rho)
\end{array} \right.
% \nonumber min\, &\, \left(-f\left(\pi,\rho\right),\varphi\left(\rho\right)\right)\\
\nonumber s.t.\, & \, \varphi\left(\rho\right) \leq C
\end{flalign}

As a bi-objective optimisation problem, {\BOTTP} asks for a set of Pareto-optimal solutions where each feasible solution cannot be improved in a second objective without degrading quality of the first one, and vice versa. In other words, the goal is to find a set of all non-dominated feasible solutions $X\subseteq\Pi\times P$ such that for any solution $\left(\pi,\rho\right) \in X$ there is no solution $\left(\pi',\rho'\right) \in X$ such that either $\left(f\left(\pi,\rho\right)<f\left(\pi',\rho'\right)\right) \wedge \left(\varphi\left(\rho\right)\geq \varphi\left(\rho'\right)\right)$ or $\left(f\left(\pi,\rho\right)\geq f\left(\pi',\rho'\right)\right) \wedge \left(\varphi\left(\rho\right) < \varphi\left(\rho'\right)\right)$ holds, where $\Pi$ is a set of feasible tours and $P$ is a set of feasible packing plans.

%%%%%%%%%%%%%%%%%%%%%%%%%%%%%%%%%%%%%%%%%%%%%%%%%%%%%%%%%%%%%%%%%%%%%%%%%%%%%%%%%%%%%%%%%%%%%%%%%%%%%%%%%%%%%%%%%%
\section{Prerequisites}\label{sec:prelim}
%%%%%%%%%%%%%%%%%%%%%%%%%%%%%%%%%%%%%%%%%%%%%%%%%%%%%%%%%%%%%%%%%%%%%%%%%%%%%%%%%%%%%%%%%%%%%%%%%%%%%%%%%%%%%%%%%%

The packing while travelling problem ({\PWT}) is a special case of the {\TTP}, which maximises the total reward for a specific tour $\pi$~\cite{DBLP:journals/eor/PolyakovskiyN17}. Thus, an optimal solution of the {\PWT} defines a subset of items producing the maximal gain. This yields a non-linear knapsack problem, which can be efficiently solved via the dynamic programming ({\DP}) approach proposed by \citet{DBLP:journals/corr/NeumannPSSW17}. Most importantly, we find that the DP yields not just a single optimal packing plan, but a set of plans $\overbar{P}_{\pi} \subseteq P$, where $(\pi, \rho)$ and $(\pi, \rho')$ do not dominate each other for any $\rho, \rho' \in \overbar{P}_{\pi}$. We name the corresponding objective vectors of $\overbar{P}_{\pi}$ as a \emph{DP front}. In Section~\ref{sec:approach}, we design our hybrid algorithm that takes advantage of the features of a DP front. 

For self-sufficiency of the paper, in Section~\ref{sec:dp}, we first briefly explain the {\DP} and how we adopt it to obtain a {\DP} front. Section \ref{sec:init} then discusses several algorithms to obtain tours that are later utilised by the {\DP} to create multiple DP fronts and to initialise the population for our hybrid evolutionary approach.

\subsection{Dynamic Programming for the {\PWT} }\label{sec:dp}

The {\DP} for the {\PWT} bases on a scheme traditional to the classical 0-1 knapsack problem. It processes items in the lexicographic order as they appear along a given tour $\pi$; that is, item $l \in \pi_i$ strictly precedes item $k \in \pi_j$, to be written as $l \preceq k$, if either $\pi_i<\pi_j$ or $\left(\pi_i=\pi_j\right) \wedge \left(l \leq k\right)$ holds. Its table $B$ is an $m \times C$ matrix, where entry $\beta_{kw}$ represents the maximal reward that can be achieved by examining all combinations of items $l$ with $l \preceq k$ leading to the weight equal to $w$. The base case of the {\DP} with respect to the first item $k$, according to the precedence order, positioned in node $\pi_i$ is as follows:
$$\beta_{kw} =
\begin{cases}
-\frac{R}{\upsilon_{max}} \left( \displaystyle\sum_{j = 1}^{n-1}d_{\pi_j \pi_{j+1}}\!+\!d_{\pi_n \pi_1} \right),& \mbox{if } w=0
\\
p_{\pi_ik} - R \left(\displaystyle\sum_{j = 1}^{n-1}\frac{d_{\pi_j \pi_{j+1}}}{\upsilon_{\pi_j}}\!+\! \frac{d_{\pi_n \pi_1}}{\upsilon_{\pi_n}}\right),& \mbox{if } w=w_{\pi_ik}
\\
-\infty,& \mbox{if } w\notin\left\{0,w_{\pi_ik}\right\}
\end{cases}
$$ 
Here, the first case relates to the empty packing when the thief collects no items at all while travelling along $\pi$, and the second computes the reward when only item $k$ is chosen. Where a combination yielding $w$ doesn't exist, $\beta_{kw}=-\infty$. For the general case, let item $l$ be the predecessor of item $k$ with regard to the precedence order. And let $\beta(k\cdot)$ denote the column containing all the entries $\beta_{kw}$ for $w\in\left[0,C\right]$. Then based on $\beta(l\cdot)$ one can obtain $\beta(k\cdot)$ computing each entry $\beta_{kw}$, assuming that item $k$ is in node $\pi_i$, as
$$max
\begin{cases}
\beta_{lw}
\\
\beta_{lw-w_{\pi_ik}} \!+\! p_{\pi_ik} \!-\! R\displaystyle\sum_{j=i}^{n-1} \left(\frac{d_{\pi_j\pi_{j+1}}}{v_{max}-\nu w} \!-\! \frac{d_{\pi_j\pi_{j+1}}}{v_{max}-\nu( w - w_{\pi_ik})}\right) 
\\
\,\,\,\,\,\,\,\,\,\,- R\left(\frac{d_{\pi_n\pi_{1}}}{v_{max}-\nu w} \!-\! \frac{d_{\pi_n\pi_{1}}}{v_{max}-\nu( w - w_{\pi_ik})}\right)
\end{cases}
$$ 
In order to reduce the search space, in each column the cells dominated by other cells are to be eliminated, i.e. if $\beta_{kw_1} > \beta_{kw_2}$ and $w_1 \leq w_2$, then $\beta_{kw_2}=-\infty$. An optimal solution derived by the {\DP} corresponds to the maximal reward stored in the last column of $B$. That is, $\max_w \left\{\beta(s,w)\right\}$ is the value of an optimal solution, where $s$ is the last item according to the precedence order.

The last column of $B$ can be considered as a complete set of non-dominated packing plans $\overbar{P}_{\pi} \subseteq P_{\pi} \subseteq P$, where $P_{\pi}$ is the set of all feasible packing plans for a given tour $\pi$. The packing plans in $\overbar{P}_{\pi}$ are non-dominated exclusion of any dominated solutions during the solution construction process.

\begin{definition} Letting $\tau$ and $T_{\pi}$ be the corresponding objective vectors sets of $\overbar{P}_{\pi}$ and $P_{\pi}$ respectively, $\tau$ is the Pareto front of $T_{\pi}$. We therefore name $\tau$ as a \emph{DP front} for the given tour $\pi$.
\end{definition}

A DP front $\tau$ for a tour $\pi$ is a complete non-dominated set, as it contains all non-dominated objective vectors in $T_{\pi}$. We take advantage of this completeness to generate the spread of solutions in our bi-objective approach in Section~\ref{sec:approach}.

\subsection{Generation of Multiple DP Fronts}\label{sec:init}
As a single DP front $\tau$ is produced for a single given tour $\pi$, i.e. $\pi \mapsto \tau$, we could generate multiple TSP tours to get a set of DP fronts.
In practice, various algorithms are capable of producing superior tours for the {\TSP}, and therefore many approaches to the {\TTP} use this capability to succeed. High-performing {\TTP} algorithms are commonly two-stage heuristic approaches, like those proposed by \citet{DBLP:conf/gecco/PolyakovskiyB0MN14}, \citet{Faulkner2015}, and \citet{ElYafrani:2016:PVS:2908812.2908847}. Specifically, their first step generates a near-optimal TSP tour and the second step completes solution by selection of a subset of items. Most of the approaches utilise the Chained Lin-Kernighan heuristic~\cite{DBLP:journals/informs/ApplegateCR03}, because it is able to provide very tight upper bounds for {\TSP} instances in short time. The knapsack component then is often handled via constructive heuristics or evolutionary approaches. However, the {\TTP} is essentially structured in such way that the importance of its both components is almost equal within the problem. Although near-optimal {\TSP} solutions can give good solutions to the {\TTP}, most of them are far away from being optimal~\cite{DBLP:conf/seal/Wu0PN17}. This is the reason for our first experimental study here, where we investigate the impact of several {\TSP} algorithms on {\TTP} solutions. Note that owing to the {\DP} we are able to solve the knapsack part to optimality, which contributes to the validity of our findings.

We analysed five algorithms for the {\TSP}: the Inver-over heuristic (INV)~\cite{DBLP:conf/ppsn/TaoM98}, the exact solver Concorde (CON)~\cite{cook2005concorde}, the ant colony-based approach (ACO)~\cite{DBLP:books/daglib/0013523}, the Chained Lin-Kernighan heuristic (LKH)~\cite{DBLP:journals/informs/ApplegateCR03} and its latest implementation (LKH2)~\cite{DBLP:journals/eor/Helsgaun00}. We ran each algorithm $10,000$ times on every instance of the \textit{eil76} series of the {\TTP} benchmark suite \cite{DBLP:conf/gecco/PolyakovskiyB0MN14}. We computed $100$ (capped due to practical reasons) distinct tours by INV, $25$ by CON, $24$ by the both ACO and LKH, and $12$ by LKH2. The lengths of the tours generated by INV are narrowly distributed around the average of $588.64$ with the standard deviation being $2.55$. By contrast, every other algorithm generates tours having the identical tour length of $585$, which beats INV.

\begin{figure}[htbp]
\centering
\includegraphics[width=0.9\linewidth]{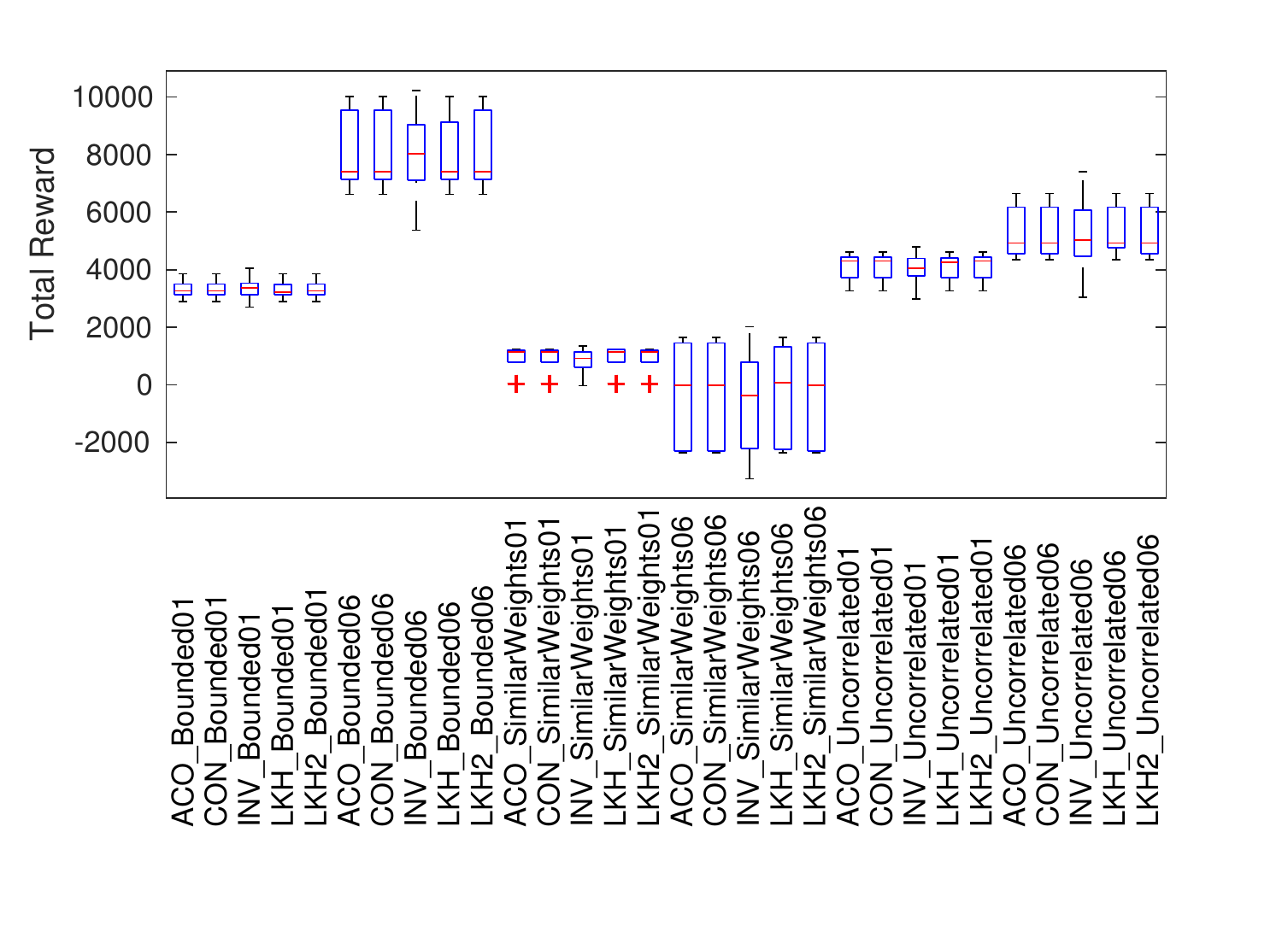}
\caption{Exploring diversity of {\TSP} tours on the \textit{eil76\_n75} series of the {\TTP} instances.}
\label{fig:initalgo}
\end{figure}

We then applied the {\DP} to every tour produced by each of the algorithms. Figure~\ref{fig:initalgo} depicts the resulted rewards on some sample TTP instances, where each box with whiskers reports the distribution of the rewards for a certain instance and the corresponding algorithm. The central mark of each box indicates the median of rewards, and the bottom and top edges of the box indicate the 25th and 75th percentiles, respectively. The whiskers extend to the most extreme rewards without considering outliers, and outliers are plotted individually as plus signs.
From the plot, we may observe that the tours generated by the CON, ACO, LKH and LKH2 have similar distributions of rewards. By contrast, the boxes of INV seem to be more extreme on the both sides. This means that the distribution of rewards via INV is more diverse and the best of the rewards outperform the others. In other words, though the Inver-over heuristic may lose against modern {\TSP} approaches, it performs better in the role of generator of varied tours for the {\TTP}. It may act as a seeding algorithm for a population in evolutionary algorithms.

\begin{figure}[htbp]
     \centering
     \includegraphics[width=0.9\linewidth]{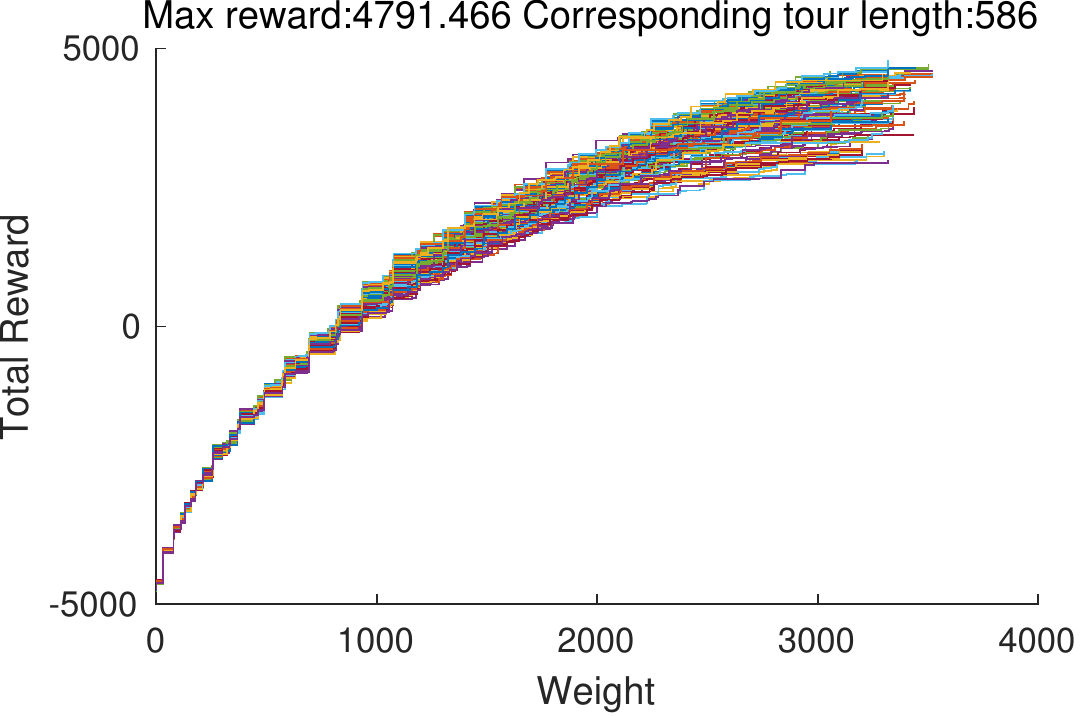}
	 \caption{The visualisation of 200 DP fronts, generated according to 100 TSP tours produced by Inver-over for the {\TTP} instance \textit{eil76\_n75\_uncorr\_01}.}
	 \label{fig:inv}
\end{figure}

In Figure~\ref{fig:inv}, we visualise the collection of the DP fronts produced by the DP on the {\TTP} instance \textit{eil76\_n75\_uncorr\_01}~\cite{DBLP:conf/gecco/PolyakovskiyB0MN14}. The corresponding tours are the $100$ tours generated by the Inver-over heuristic. Actually, the plot depicts $200$ fronts since the {\DP} was applied to a tour and its reversed order.

\begin{definition} Given $n$ DP fronts $\tau_1,\dots,\tau_n$, let $\Phi$ denote a union of the fronts as $\Phi=\cup_{i=1}^n \tau_i$. Then a subset $\omega \subseteq \Phi$ is the Pareto front of $\Phi$ called as the \emph{surface} of $\Phi$. 
\end{definition}\label{def:surface}

The surface $\omega$ is formed by the union of all superior points resulted from different DP fronts in $\Phi$. It is further used to guide evolution process in our approach.
 
%%%%%%%%%%%%%%%%%%%%%%%%%%%%%%%%%%%%%%%%%%%%%%%%%%%%%%%%%%%%%%%%%%%%%%%%%%%%%%%%%%%%%%%%%%%%%%%%%%%%%%%%%%%%%%%%%%
\section{A hybrid evolutionary approach}\label{sec:approach}
%%%%%%%%%%%%%%%%%%%%%%%%%%%%%%%%%%%%%%%%%%%%%%%%%%%%%%%%%%%%%%%%%%%%%%%%%%%%%%%%%%%%%%%%%%%%%%%%%%%%%%%%%%%%%%%%%%

Multi-objective optimisation algorithms guided by evolutionary mechanisms explore the decision space iteratively in order to determine a set of Pareto-optimal solutions. Indeed, many of them may act myopically as they sample the space searching for individual solutions without clear vision of the whole picture in terms of other solutions and their number. Therefore, achieving strong diversity in exploring the space plays an important role in evolutionary algorithm design. In this paper, we discuss one way to overcome potential issues related to diversity and propose a hybrid approach where evolutionary techniques and dynamic programming find synergy in their combination. 

One of the challenges of multi-objective optimisation is to keep the wide spread of solutions, which has to be guaranteed by strong diversity. Modern approaches normally incorporate additional processes to tackle this, such as the density estimation and/or crowdedness-comparison operator in SPEA2~\cite{Zitzler01spea2} and NSGA-II~\cite{DBLP:journals/tec/DebAPM02}. %, which are computationally costly. 
In our approach, the {\DP} is incorporated as a subroutine capable of producing at once a series of possible decisions with regard to a given tour. Thus, when a tour is specified, the {\DP} guarantees that a corresponding front will be built without missing any of its points due to the completeness of the DP front, which thus also guarantees a good spread of solutions.

On the other hand, due to the typically observed non-dominance of single DP fronts, the global Pareto optimality of the BO-TTP may be formed either by a single DP front or by the combination of segments from different top DP fronts. In Figure~\ref{fig:inv}, we may observe that the DP fronts are all intertwined together, including the ones at the surface of the fronts collection. This seems to indicate that the Pareto-optimal set of solutions is more likely to be the result of multiple TSP tours and their DP fronts. We would like our evolutionary mechanism to take advantage of this and to keep the top DP fronts so as to improve the population further. In order to achieve this as well as to overcome the drawback of existing multi-objective evolutionary optimisation algorithms that focus on individual solutions, we design our hybrid IBEA with particular indicators and selection mechanisms in orchestrating improvement of Pareto front guided by the information of the DP fronts for most promising {\TSP} tours.

Our hybrid approach reduces the search space to some extent by decomposing the problem and thus transforming it. Evolutionary optimisation approaches traditionally depend on the choice of solution encoding (i.e. chromosome). %Although encoding may involve entirely the same decision variables as those in an original problem, some approaches can alter the problem to gain in algorithm design or improve efficiency and performance. 
Our approach %proceeds similarly to those techniques and 
treats a single {\TSP} tour as an individual. Thus, a set of tours yields a population. Indeed, it operates on a reduced set of variables (implying shorter chromosomes), thus decreasing memory consumption and the number of internally needed sorting operations, comparisons and search operations. 
%As it does not explicitly include binary variables required for the packing part.

\begin{algorithm}[htbp]
\caption{Hybrid IBEA Approach}
\label{alg:ibea}
{\small
\begin{algorithmic}
\State \textbf{Input:} population size $\mu$; limit on the number of generations $\alpha$;
\State \textbf{Initialisation:} 
\State set the iteration counter $c = 0$;
\State populate $\overbar{\Pi}$ with $\mu$ new tours produced by the TSP solver; 
\While{($c \leq \alpha$)}
	\State set $c=c+1$;
	\State \textbf{Indicator:}
	\State run the {\DP} for every tour $\pi \in \overbar{\Pi}$ to compute its DP front $\tau$;
	\State apply indicator function $\mathcal{I}(\tau)$ to calculate the indicator value for every individual tour $\pi \in \overbar{\Pi}$;
	\State \textbf{Survivor Selection:}
	\State repeatedly remove the individual with the smallest indicator value from the population $\overbar{\Pi}$ until the population size is $\mu$ (ties are broken randomly);
	\State \textbf{Parent Selection:}
	\State apply parent selection procedure to $\overbar{\Pi}$ according to the indicator values to choose a set $\Lambda$ of $\lambda$ parent individuals;
	\State \textbf{Mating:}
	\State apply crossover and mutation operators to the parents of $\Lambda$ to obtain a child population $\Lambda'$;
	\State set the new population as $\overbar{\Pi} = \overbar{\Pi} \cup \Lambda'$;
\EndWhile
\end{algorithmic}}
\end{algorithm}

Algorithm~\ref{alg:ibea} sketches the whole approach, which we adopted from the original IBEA introduced by \citet{DBLP:conf/ppsn/ZitzlerK04}.
It accepts $\mu$ as a control parameter for the size of the population $\overbar{\Pi} \subseteq \Pi$ and $\alpha$ as a limit on the number of iterations, which defines its termination criterion. In order to utilise the information within the DP fronts to guide the evolution of individual tours, we design new indicators to be computed based on the DP fronts instead of directly on the individuals. Our specific selection mechanisms then filter the individuals according to the indicator values in order to find the tours with better DP fronts.

The rest of this section first introduces the indicator functions we apply to {\TSP} solutions. Next, it details a parent selection mechanism to mate existing individuals from the population. It ends with a discussion of mutation and crossover operators guiding the search.

%%%%%%%%%%%%%%%%%%%%%%%%%%%%%%%%%%%%%%%%%%%%%%%%%%%%%%%%%%%%%%%%%%%%%%%%%%%%%%%%%%%%%%%%%%%%%%%%%%%%%%%%%%%%%%%%%%
\subsection{Design of Indicators}\label{sec:indicator}
%%%%%%%%%%%%%%%%%%%%%%%%%%%%%%%%%%%%%%%%%%%%%%%%%%%%%%%%%%%%%%%%%%%%%%%%%%%%%%%%%%%%%%%%%%%%%%%%%%%%%%%%%%%%%%%%%%
The designs of our indicators are based on the idea of measuring how each DP front contributes to the surface $\omega$ of the fronts' union $\Phi$ corresponding to the population $\overbar{\Pi}$. The surface $\omega$ introduced in Definition~\ref{def:surface} is the union of all best segments from different DP fronts in $\Phi$. 
Given a DP front $\tau$ for a tour $\pi \in \overbar{\Pi}$ and a measurement function $\mathcal{M}$ of a front, we use the followed formula to calculate the indicator $\mathcal{I}$:
\begin{align}
\mathcal{I}(\tau)= 1- \frac{\mathcal{M}(\omega \setminus \tau)}{\mathcal{M}(\omega)}. \label{fml:ind}
\end{align}
This formula measures how much we could lose (expressed as a value from 0 to 1) if we did not include the segments of the front $\tau$ to the surface $\omega$, i.e. $\omega \setminus \tau$.
In the following, we study two types of the measurement functions: Surface Contribution (SC) and Hypervolume (HV), hence two corresponding indicators: the Loss of Surface Contribution (LSC) and the Loss of Hypervolume (LHV).

\textbf{Loss of Surface Contribution}. 
Our first indicator is Surface Contribution (SC), which is a novel and direct measure. Given the union of a set of fronts $\Phi$, a front $\tau \subseteq \Phi$ and the surface $\omega \subseteq \Phi$, $SC(\tau)$ counts the number of objective vectors that $\tau$ contributes to $\omega$, as defined by: 
\begin{align}
 SC(\tau) = \frac{|\omega \cap \tau|}{|\omega|}. \label{fml:sc}
\end{align}
 Using SC~(\ref{fml:sc}) to replace the $\mathcal{M}$ function in (\ref{fml:ind}), we have the formula of LSC as follows:
$$
 LSC(\tau) = 1 - SC(\omega \setminus \tau).
$$

\textbf{Loss of Hypervolume}. In multi-objective optimisation, the hypervolume indicator is a traditional indicator used to indicate the quality of a set of objective vectors~\cite{DBLP:conf/ppsn/ZitzlerT98}. In the bi-criteria case, when a front is given as a set of points in two-dimensional space, its value is computed as a sum of areas of rectangular regions. 

%Let $\left\{0, C\right\}$ 
Let $(0, C)$ be the reference point for our problem, which implies that only the range of non-negative objective values is taken into account. In addition, let $p=(u, v) \in \tau$ be a bi-dimensional objective vector in a DP front $\tau$ while $u>0$ and $v<C$, $HV(\tau)$ calculates the hypervolume for $\tau$ as:
$$HV(\tau)=\displaystyle\sum_{p \in \tau} u_p\left(v_{p}-v_{p-1}\right)$$
Putting $HV(\tau)$ back to (\ref{fml:ind}), we have the loss of hypervolume $LHV\left(\tau\right)$ computed as 
$$LHV\left(\tau\right)=1-\frac{HV(\omega \setminus \tau)}{HV(\omega)}.$$

%%%%%%%%%%%%%%%%%%%%%%%%%%%%%%%%%%%%%%%%%%%%%%%%%%%%%%%%%%%%%%%%%%%%%%%%%%%%%%%%%%%%%%%%%%%%%%%%%%%%%%%%%%%%%%%%%%
\subsection{Parent Selection Mechanisms}\label{sec:selection}
%%%%%%%%%%%%%%%%%%%%%%%%%%%%%%%%%%%%%%%%%%%%%%%%%%%%%%%%%%%%%%%%%%%%%%%%%%%%%%%%%%%%%%%%%%%%%%%%%%%%%%%%%%%%%%%%%%
With the individuals in the population $\overbar{\Pi}$ being measured by the defined indicators, we can study strategies that shall efficiently select good individuals. There are five parent selection schemes that we take into consideration due to their popularity or previous theoretical findings. In comparison, we introduce two simple and arbitrary selections as well as a traditional policy to be a baseline. In this study, we expect to find a well-performing combination of indicator and selection to encourage the synergy of the DP and evolutionary approach. 
  
\textbf{Rank-based Selection} (RBS). In the rank-based selection policy, individuals are first ranked with respect to the value of an indicator. The selection policy is based then on a specific distribution law affecting the choice of a parent. Here, we study three schemes introduced by~\citet{DBLP:conf/gecco/OsunaGNS17}, namely exponential (EXP), inverse quadratic (IQ) and Harmonic (HAR), and make them a part of our hybrid approach. Given a population of size $\mu$, the probability of selecting the $i$th ranked individual according to EXP, IQ and HAR is, respectively,
 
 \begin{align}
 	\frac{2^{-i}}{\sum_{j=1}^\mu{2^{-j}}},~
	\frac{i^{-2}}{\sum_{j=1}^\mu{j^-{2}}},~
	\frac{i^{-1}}{\sum_{j=1}^\mu{j^{-1}}}.
 \end{align}
 
\textbf{Fitness-Proportionate Selection} (FPS). This rule estimates an individual $\pi \in \overbar{\Pi}$ according to the indicator $\mathcal{I}(\tau)$ of its DP front $\tau$. It has the following form:
 \begin{align}
FPS\left(\pi_i\right) = \frac{\mathcal{I}(\tau_i)}{\sum_{j=1}^\mu{\mathcal{I}(\tau_j)}}.
 \end{align}
 
\textbf{Tournament Selection} (TS). This policy applies the tournament selection~\cite{DBLP:journals/compsys/MillerG95}, but employs indicators discussed in Section~\ref{sec:indicator} to rank individuals.
 
\textbf{Arbitrary Selection} (AS). Here, we consider two different rules: the best arbitrary selection (BST) and another one, which we call extreme (EXT). The former ranks individuals of a population with accordance to the value of an indicator and selects the best half of the population. The latter proceeds similarly selecting $25\%$ of the best and $25\%$ of the worst individuals.
 
\textbf{Uniformly-at-random Selection} (UAR). This traditional policy selects a parent from a population with probability $\frac{1}{\mu}$ uniformly at random.
 
%%%%%%%%%%%%%%%%%%%%%%%%%%%%%%%%%%%%%%%%%%%%%%%%%%%%%%%%%%%%%%%%%%%%%%%%%%%%%%%%%%%%%%%%%%%%%%%%%%%%%%%%%%%%%%%%%%
\subsection{Mutation and Crossover Operators}
%%%%%%%%%%%%%%%%%%%%%%%%%%%%%%%%%%%%%%%%%%%%%%%%%%%%%%%%%%%%%%%%%%%%%%%%%%%%%%%%%%%%%%%%%%%%%%%%%%%%%%%%%%%%%%%%%%
In our approach, we adopt a multi-point crossover operator that has already proved its efficiency for the {\TTP} in  \cite{ElYafrani:2016:PVS:2908812.2908847}. As an (un-optimised) rule, we perform the crossover operation on a tour with $80\%$ probability. It is always followed by the mutation procedure, which either applies the classical 2OPT mutation~\cite{Croes1958} or re-inserts a node to another location. Both the node and the location are selected uniformly at random. We name these two operators 2OPT and JUMP, respectively.

%%%%%%%%%%%%%%%%%%%%%%%%%%%%%%%%%%%%%%%%%%%%%%%%%%%%%%%%%%%%%%%%%%%%%%%%%%%%%%%%%%%%%%%%%%%%%%%%%%%%%%%%%%%%%%%%%%
\section{Computational Experiments}\label{sec:experiments}
%%%%%%%%%%%%%%%%%%%%%%%%%%%%%%%%%%%%%%%%%%%%%%%%%%%%%%%%%%%%%%%%%%%%%%%%%%%%%%%%%%%%%%%%%%%%%%%%%%%%%%%%%%%%%%%%%%

\subsection{Computational Set Up}\label{sec:setup}

We examine the IBEA presented in Algorithm~\ref{alg:ibea} by going through each of the two indicators and the eight parent selections, resulting in a total of 16 settings. %Each setting represents a combination of one of the 2 indicators and one of the 8 parent selections respectively. 
For example, FPS on LHV means the combination of the FPS selection and the LHV indicator. 

From the original set of TTP instances, we use three different types, namely bounded-strongly-correlated (\emph{Bounded}), uncorrelated (\emph{Uncorrelated}) and un-correlated-with-similar-weights (\emph{SimilarWeights}), selected from three instance series: \emph{eil51}, \emph{eil76}, \emph{eil101} in the TTP benchmark~\cite{DBLP:conf/gecco/PolyakovskiyB0MN14}. We run our approach 30 times repetitively on each selected instance. Each time, the algorithm runs 20,000 generations on a population $\overbar{\Pi}$ in size of 50.

Due to the significant computing cost, our experiments run on the supercomputer in our university, which consists of 5568 Intel(R) Xeon(R) 2.30GHz CPU cores and 12TB of memory. %We allocate 16 CPU cores and 16GB of memory to each individual experiment. Overall, 69,120 CPU cores and the same number of GB memory have been allocated for the experiments. 
Overall the experiments consumed around 170,000 CPU-hours.

 %%%%%%%%%%%%%%%%%%%%%%%%%%%%%%%%%%%%%%%%%%%%%%%%%%%%%%%%%%%%%%%%%%%%%%%%%%%%%%%%%%%%%%%%%%%%%%%%%%%%%%%%%%%%%%%%%%
 \subsection{Results and Analysis}\label{sec:results}
 %%%%%%%%%%%%%%%%%%%%%%%%%%%%%%%%%%%%%%%%%%%%%%%%%%%%%%%%%%%%%%%%%%%%%%%%%%%%%%%%%%%%%%%%%%%%%%%%%%%%%%%%%%%%%%%%%%
%In order to evaluate the population $\overbar{\Pi}$ to concisely present our results, we calculate the hypervolume of its surface $\omega$. Thus we have the mean and standard deviation (SD) of the hypervolumes of $30$ independent runs for each test case. We also store the corresponding total reward in order to compare with the results from the state-of-the-art single-objective approach: MA2B~\cite{ElYafrani:2016:PVS:2908812.2908847}.
To compare the outcomes of the different approaches based on the final populations $\overbar{\Pi}$ of tours, we calculate the hypervolumes for the surface of resulting non-dominated solutions. 
We also store the corresponding total reward in order to compare with the results from the state-of-the-art single-objective approach: MA2B~\cite{ElYafrani:2016:PVS:2908812.2908847} (see comparison in~\cite{Wagner2017ttpalgssel}).

However, due to the varied mean values and unknown global optima of different TTP instances, it is hard to analyse and compare across instances. Nevertheless, such a comparison is desired because such analysis or comparison may provide a more general view for our algorithm. We design a statistical comparison to overcome this as follows. Firstly, we choose the uniformly-at-random (UAR) selection as the baseline, which creates two baseline settings, namely UAR on LHV and UAR on LSC. We secondly conduct \emph{Welch's t-test}~\cite{welch1947} between the results of the others and the baselines for two indicators respectively.

The results of the t-test are probability values (p-values), each of which measures the likelihood of one selection to the corresponding baseline with respect to their performance. For example, we have the p-value being $4.75 \times 10^{-7}$ in the case of comparing the hypervolume of the FPS and the UAR on LHV. This means that the probability of the FPS performing identical to the UAR on LHV (as expressed by having the same means) is less than $0.0000475\%$. In fact, the former performs much better than the latter on average. In order to improve the readability, we use the logarithm of the p-value in our plots. Thus, the measure of the FPS on LHV in our little example is $6.32$ (i.e. $\log_{10}{(4.75 \times 10^{-7})}$). In short, the larger the logarithmic p-value is, the better the selection is against the UAR.

 \begin{figure}[htbp]
     \centering
	 \begin{subfigure}[b]{\linewidth}
         \includegraphics[width=\linewidth]{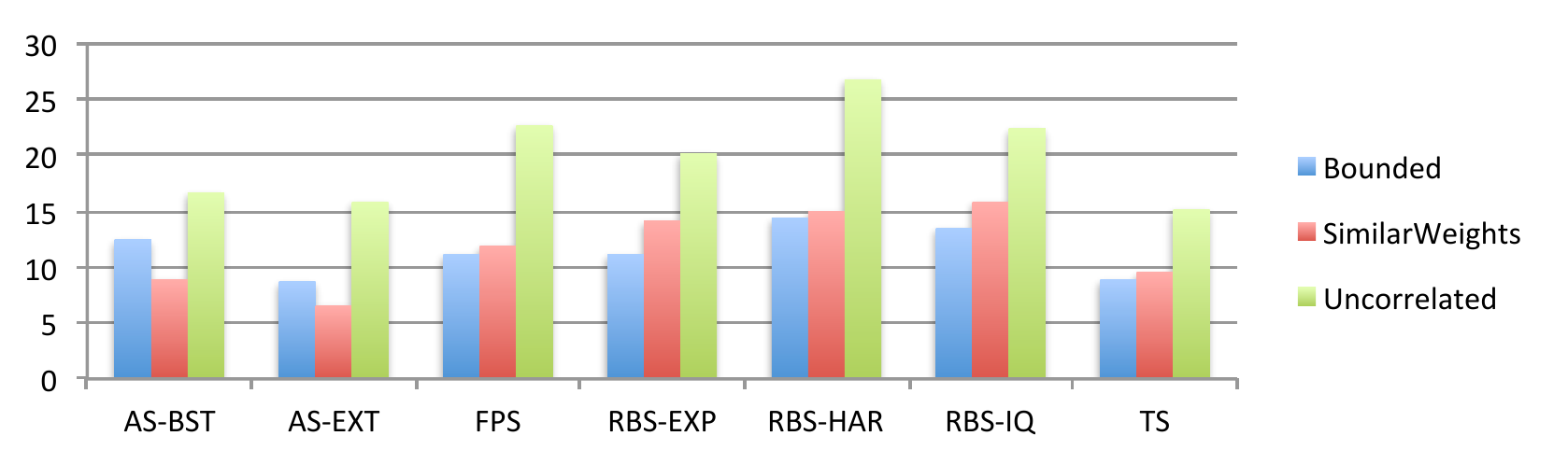}
		 \caption{Hypervolume of each selection on LHV}
     \end{subfigure}
	 \begin{subfigure}[b]{\linewidth}
         \includegraphics[width=\linewidth]{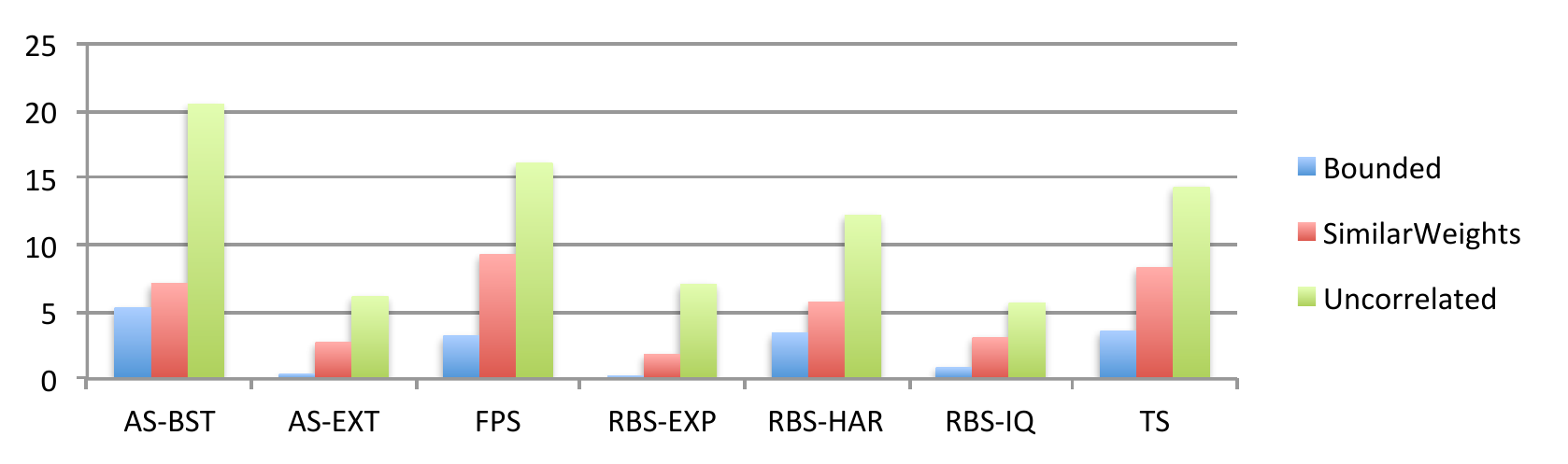}
		 \caption{Hypervolume of each selection on LSC}
     \end{subfigure}
	 
	 \begin{subfigure}[b]{\linewidth}
         \includegraphics[width=\linewidth]{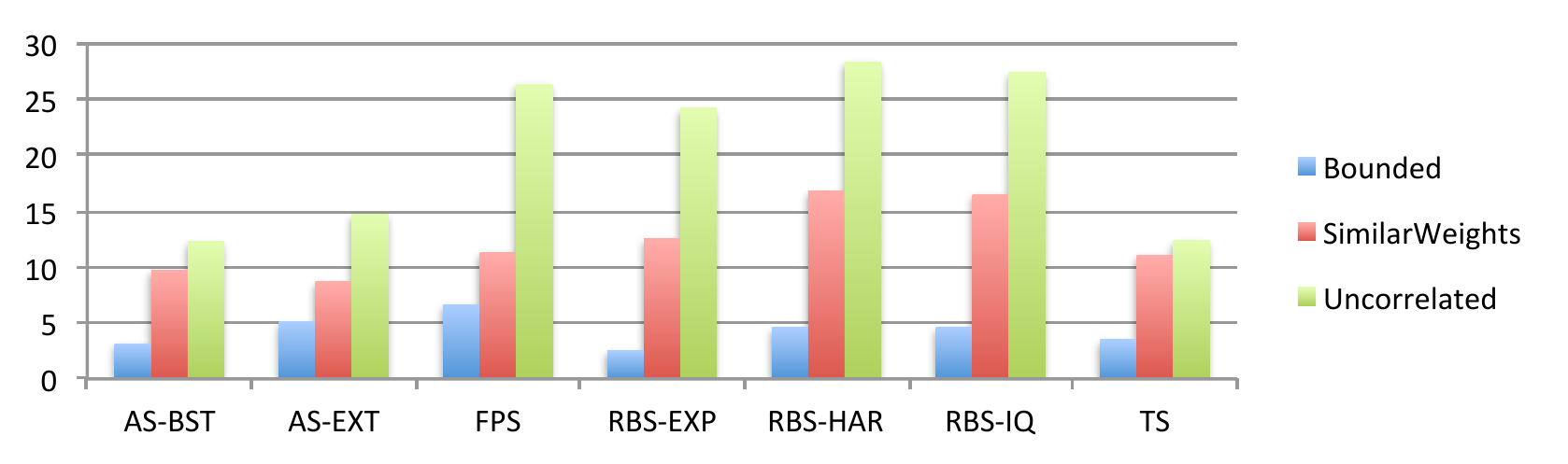}
		 \caption{Total Reward of each selection on LHV}
     \end{subfigure}
	 \begin{subfigure}[b]{\linewidth}
         \includegraphics[width=\linewidth]{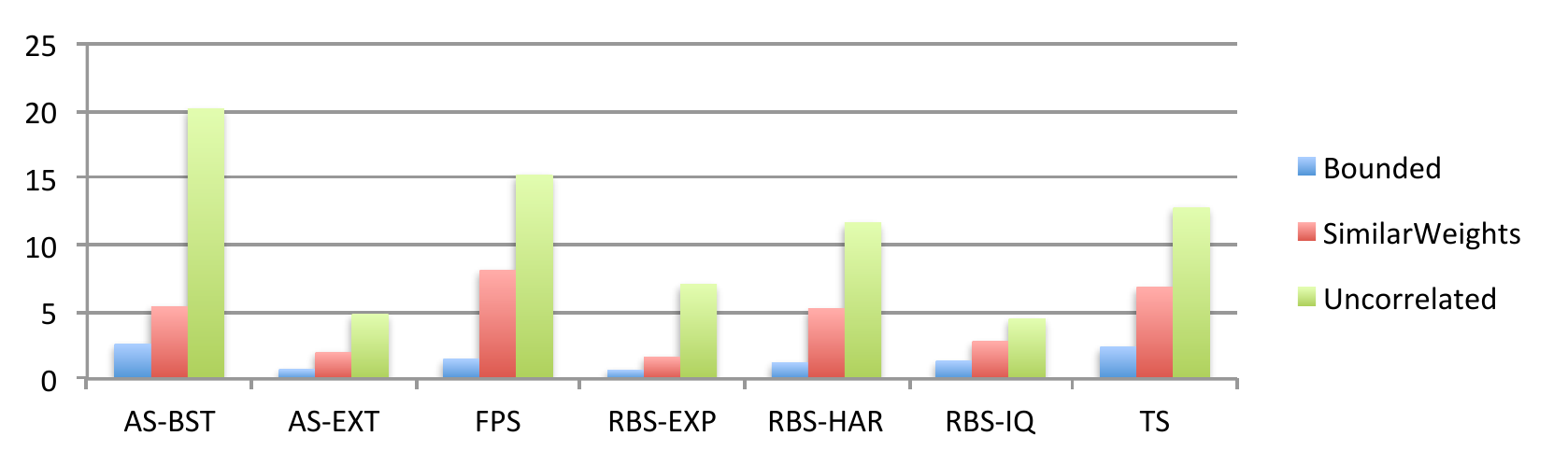}
		 \caption{Total Reward of each selection on LSC}
     \end{subfigure}

	 \caption{The sum of the logarithm of the p-values of performing Welch's t-test for the selections each respectively against the UAR selection.}
	 \label{fig:reward}
\end{figure}

Figure~\ref{fig:reward} depicts the overall results of the Welch's t-test, in which we categorise our results into three types of bars according to three types of TTP instances: Bounded, Uncorrelated and SimilarWeights. Each bar in the plots represents the mean of the logarithmic p-values of several instances in this category, for example \emph{eil51\_n50\_bounded-strongly-corr\_01.ttp}, \emph{eil76\_n75\_bounded-strongly-corr\_01.ttp} and \emph{eil101\_n100\_bounded-strongly-corr\_01.ttp}. From it we may observe distinguishable patterns between the selections running on the LHV and the LSC respectively. For example, the three rank-based selection (RBS) schemes generally perform better on LHV than on LSC, among which the HAR is the best. According to the definitions, the HAR is the least aggressive scheme among the three, with a fat tail and relatively small probability for selecting the best few individuals~\cite{DBLP:conf/gecco/OsunaGNS17}. It seems to imply that the LHV benefits more from the diversity of candidates. By contrast, the AS-BST performs best on LSC, which might imply that the LSC relies more on a few outstanding individuals for approximating, as the AS-BST only focuses on the best ones.

In terms of different types of TTP instances, we may observe that the IBEA performs best on the uncorrelated instances in all of the settings, while being worst on the strongly bounded ones in most of the settings. This to some extent supports the conjecture that strongly bounded TTP instances are the (relatively) hard ones and uncorrelated instances are the easy ones~\cite{DBLP:conf/gecco/PolyakovskiyB0MN14}.

With regard to the choice of the parent selections, besides the RBS-HAR and the AS-BST which perform best on LHV and LSC respectively, we would like to recommend the FPS as well. This selection seems to be the safest choice, as it performs consistently well on different settings.

\begin{table}[thbp]
\centering
{
\begin{tabular}{|ll|lll|}

\hline
\multicolumn{5}{|c|}{MA2B} \\
\hline
&&	Mean	&	Max	&	SD	\\
eil51\_n50	&	Uncorrelated	&	2805.000	&	2855	&	27.814	\\
	&	SimilarWeights	&	1416.348	&	1460	&	47.906	\\
	&	Bounded	&	4057.652	&	4105	&	25.841	\\
eil76\_n75	&	Uncorrelated	&	5275.067	&	5423	&	78.138	\\
	&	SimilarWeights	&	1398.867	&	1502	&	55.448	\\
	&	Bounded	&	3849.067	&	4109	&	139.742	\\
eil101\_n100	&	Uncorrelated	&	3339.600	&	3789	&	388.360	\\
	&	SimilarWeights	&	2215.500	&	2483	&	235.905	\\
	&	Bounded	&	4949.000	&	5137	&	139.285	\\
 \hline \hline
\multicolumn{5}{|c|}{FPS LHV} \\
 \hline
 &&	Mean	&	Max	&	SD	\\
eil51\_n50	&	Uncorrelated	&	\cellbluelight2828.728	&	2854.543	&	15.357	\\
	&	SimilarWeights	&	1413.044	&	1459.953	&	17.780	\\
	&	Bounded	&	\cellbluelight4229.149	&	\cellbluelight4230.997	&	10.118	\\
eil76\_n75	&	Uncorrelated	&	\cellbluelight5445.624	&	\cellbluelight5514.666	&	58.992	\\
	&	SimilarWeights	&	\cellbluelight1477.680	&	\cellbluelight1513.404	&	24.494	\\
	&	Bounded	&	\cellbluelight4042.449	&	4108.760	&	38.805	\\
eil101\_n100	&	Uncorrelated	&	\cellbluelight3620.844	&	\cellbluelight3943.425	&	222.815	\\
	&	SimilarWeights	&	\cellbluelight2431.907	&	2482.462	&	52.265	\\
	&	Bounded	&	\cellbluelight5094.246	&	\cellbluelight5233.513	&	65.267	\\
  \hline \hline
\multicolumn{5}{|c|}{FPS LSC} \\
  \hline
  &&	Mean	&	Max	&	SD	\\
eil51\_n50	&	Uncorrelated	&	\cellbluelight2810.509	&	2832.496	&	18.076	\\
	&	SimilarWeights	&	\cellbluelight1426.135	&	1459.953	&	21.990	\\
	&	Bounded	&	\cellbluelight4231.299	&	\cellbluelight4241.199	&	1.881	\\
eil76\_n75	&	Uncorrelated	&	\cellbluelight5392.575	&	\cellbluelight5514.666	&	73.029	\\
	&	SimilarWeights	&	\cellbluelight1474.803	&	\cellbluelight1513.404	&	21.346	\\
	&	Bounded	&	\cellbluelight4054.815	&	4102.167	&	21.440	\\
eil101\_n100	&	Uncorrelated	&	\cellbluelight3664.369	&	\cellbluelight3846.172	&	124.994	\\
	&	SimilarWeights	&	\cellbluelight2436.374	&	2482.462	&	49.731	\\
	&	Bounded	&	\cellbluelight5067.070	&	\cellbluelight5233.513	&	55.587	\\
   \hline
 \end{tabular}
 }
 \caption{Comparison of the total reward between running the state-of-art approach MA2B and the IBEA with the selection being FPS and the indicator being LHV and LSC respectively. Each approach runs $30$ times on the TTP instances. Highlighted are the results that are better than MA2B.}
\label{tab:comp}
\end{table}

Overall, we may observe from Figure~\ref{fig:reward} that the figures of the hypervolume generally agree with those of the total reward. This somewhat suggests that optimising the bi-objective TTP brings good results for the single objective TTP as well. Table~\ref{tab:comp} presents the total rewards we get by optimising the {\BOTTP}, in comparison with the state-of-art algorithm of the single objective TTP, namely MA2B~\cite{ElYafrani:2016:PVS:2908812.2908847}. We run the MA2B with the time limits identical to our approach. The results show that in the majority of the test cases, our approach preforms better.

%%%%%%%%%%%%%%%%%%%%%%%%%%%%%%%%%%%%%%%%%%%%%%%%%%%%%%%%%%%%%%%%%%%%%%%%%%%%%%%%%%%%%%%%%%%%%%%%%%%%%%%%%%%%%%%%%%
\section{Conclusion}\label{sec:conclusion}
%%%%%%%%%%%%%%%%%%%%%%%%%%%%%%%%%%%%%%%%%%%%%%%%%%%%%%%%%%%%%%%%%%%%%%%%%%%%%%%%%%%%%%%%%%%%%%%%%%%%%%%%%%%%%%%%%%
In this paper, we investigated a new bi-objective travelling thief problem which optimises both the total reward and the total weight. We proposed a hybrid indicator-based evolutionary algorithm (IBEA) that utilises the exact dynamic programming algorithm for the underlying PWT problem as a subroutine to evolve the individuals. This approach guarantees the spread of solutions without introducing additional spread mechanisms. We furthermore designed and studied novel indicators and selection schemes that take advantage of the information in the Pareto fronts generated by the exact approach for evolving solutions towards the global Pareto optimality. Our results show that this approach solves the problem well, because its by-products, which are the results for the single-objective travelling thief problem, beat the state-of-the-art approach single-objective approaches.

%%%%%%%%%%%%%%%%%%%%%%%%%%%%%%%%%%%%%%%%%%%%%%%%%%%%%%%%%%%%%%%%%%%%%%%%%%%%%%%%%%%%%%%%%%%%%%%%%%%%%%%%%%%%%%%%%%
\section*{Acknowledgements}
%%%%%%%%%%%%%%%%%%%%%%%%%%%%%%%%%%%%%%%%%%%%%%%%%%%%%%%%%%%%%%%%%%%%%%%%%%%%%%%%%%%%%%%%%%%%%%%%%%%%%%%%%%%%%%%%%%

The authors were supported by Australian Research Council grants DP130104395, DP140103400, and DE160100850.

\newpage
\balance
\bibliographystyle{ACM-Reference-Format}
\bibliography{references} 

\end{document}